\documentclass[cameraready]{Interspeech}

\usepackage{float}
\usepackage[T1]{fontenc}
\usepackage[utf8]{inputenc}
\usepackage{cite}
\usepackage{cleveref}
\usepackage{booktabs}
\usepackage{adjustbox}
\usepackage{multirow}
\usepackage{graphicx}
\usepackage{color}
\usepackage{makecell}

\title{Diffusion Language Models for Speech Recognition}

\author[affiliation={1}, equalcontribution]{Davyd}{Naveriani}
\author[affiliation={1,2}, orcid=0000-0002-6655-671X, equalcontribution]{Albert}{Zeyer}
\author[affiliation={1,2}, orcid=0000-0003-2839-9247]{Ralf}{Schlüter}
\author[affiliation={1,2}]{Hermann}{Ney}

\address{
$^1$Machine Learning and Human Language Technology Group, RWTH Aachen University, Germany \\
$^2$AppTek, Germany
}

\email{davyd.naveriani@rwth-aachen.de, \{zeyer,schlueter,ney\}@cs.rwth-aachen.de}
\keywords{Speech Recognition, Diffusion Language Model}

\graphicspath{{figures/}}

\makeatletter
\renewcommand{\paragraph}[1]{\par\addvspace{0.5em \@plus1ex \@minus.2ex}\noindent\textbf{#1}\quad\ignorespaces}
\makeatother

\begin{document}

\maketitle

\begin{abstract}
Diffusion language models have recently emerged
as a leading alternative to standard language models,
due to their ability for bidirectional attention and parallel text generation. 
In this work, we explore variants for their use in speech recognition. 
Specifically, we introduce a comprehensive guide
to incorporating masked diffusion language models (MDLM)
and uniform-state diffusion models (USDMs)
for rescoring ASR hypotheses. 
Additionally, we design a new joint-decoding method
that combines CTC and USDM
by integrating the framewise probability distributions
derived from CTC
with the labelwise probability distributions
computed by USDM at each decoding step,
thereby generating new candidates that combine strong language knowledge
from USDM and acoustic information from CTC. 
Our findings reveal that USDM, as well as MDLM,
can significantly improve the accuracy of recognized text.
We publish all our code and recipes.
\end{abstract}

\section{Introduction}
Autoregressive language models (LMs) are commonly used
to improve automatic speech recognition (ASR) systems
due to their strong linguistic capabilities
and the ability to incorporate external textual knowledge
\cite{jelinek1998statistical,irie19:trafolm,prabhavalkar2023endtoend}.
However, applying traditional autoregressive LMs in joint decoding inherently limits the speed due to their strictly left-to-right decoding structure.
Non-autoregressive language models \cite{su2021non}
can operate in a parallel manner,
potentially resolving this bottleneck and enabling faster decoding.

Recently,
discrete diffusion models,
such as large language diffusion with masking (LLaDA) \cite{nie2025large}
and masked diffusion language models (MDLM) \cite{sahoo2024simple}
have emerged as powerful non-autoregressive alternatives
\cite{sahoo2024simple,sahoo2025diffusion,nie2025large,li2025survey,ni2025diffusion,labs2025mercuryultrafastlanguagemodels}.
Prior work has explored diffusion-based models in ASR as audio-conditioned decoders \cite{wang2025audio,kwon2025whisfusion,tian2026dllmasrfasterdiffusionllmbased}.
Other non-autoregressive speech recognition models have been proposed
\cite{higuchi2021comparative, deng2022improving, navon2025draxspeechrecognitiondiscrete}.

Here, we focus on keeping a separate language model
to better leverage large amounts of text data
and to allow more flexible integration of the language model into the ASR system \cite{gulcehre2015usingmonolingualcorporaneural,toshniwal2018comparisontechniqueslanguagemodel}.
An investigation of
diffusion language models as standalone models for joint ASR decoding
has not yet been conducted
and their integration into token-level joint decoding
remains unexplored.

In this work,
we systematically investigate diffusion language models for ASR rescoring,
comparing masked diffusion language models (MDLM)
and uniform-state diffusion models (USDM).
Furthermore, we develop a novel token-level decoding method
that allows for the integration of USDM
with a non-autoregressive connectionist temporal classification (CTC)
speech recognition model \cite{10.1145/1143844.1143891}.
Because USDM corrupts sequences using uniform transitions without artificial mask tokens, 
it provides a full vocabulary probability distribution for every token at each denoising step,
which enables a direct combination of framewise CTC probabilities and labelwise diffusion distributions during hypothesis construction, 
as detailed in~\Cref{sec:diffusionlm} and~\Cref{sec:methodology}.
To the best of our knowledge,
this is the first work that (i) systematically compares masked
and uniform-state diffusion language models for ASR rescoring,
and (ii) integrates a uniform-state diffusion language model into CTC-based token-level joint decoding.

\section{Diffusion Language Models}
\label{sec:diffusionlm}

\paragraph{Masked diffusion language model.}
MDLM corrupts text by randomly masking tokens 
and learns to reconstruct the sequence during the reverse generative pass. 

During the forward process, 
tokens are independently masked based on a monotonically decreasing noise schedule $\alpha_t \in [0, 1]$. 
Essentially, $\alpha_t$ represents the probability of a token retaining its original value, 
while $(1 - \alpha_t)$ is the probability of it being masked. 
As the process reaches the final step $T$, $\alpha_T$ approaches $0$, 
meaning the sequence becomes completely masked with probability $1$. 
The marginal distribution of this forward process is defined as:
\begin{equation}
q(z_t \mid w) = \mathrm{Cat}(z_t;\, \alpha_t {1}_w + (1 - \alpha_t) {1}_m)
\label{eq:mdlm_forward}
\end{equation}
where $\mathrm{Cat}$ is the categorical distribution, $w$ is the original clean token,
$z_t \in V$ is the token at diffusion step $t$,
${1}_w \in \{0,1\}^{|V|}$ denotes the one-hot vector with a $1$ at position $w \in V$ and indicates a clean token, 
and $m$ is the index of the \texttt{[MASK]} token. 

During the reverse process, 
starting from a fully masked sequence, 
an MDLM iteratively denoises the text to recover the original tokens. 
The theoretical objective is to align the reverse transition with the true posterior distribution, 
which is defined as:
\begin{align}
  & q(z_s \mid z_t, w) \nonumber \\
  & =
  \begin{cases}
  \mathrm{Cat}(z_s; {1}_{z_t}) & z_t \neq m, \\
  \mathrm{Cat}\left(z_s; \dfrac{(1 - \alpha_s){1}_{m}+ (\alpha_s - \alpha_t){1}_{w}}{1 - \alpha_t}\right) & z_t = m.
  \end{cases}
  \label{eq:mdlm_posterior}
\end{align}
However, 
because the target token $w$ is unknown during generation, 
a parameterized model $w_\theta(z_t, t)$ is trained to directly predict the best estimate of the unmasked tokens from the noisy state $z_t$. 
This implies that the training objective is computed exclusively over the masked tokens, 
taking the form of a cross-entropy loss weighted by the noise schedule~\cite{sahoo2024simple, nie2025large}.
The bidirectional context modeling of MDLM makes it 
particularly well-suited for ASR hypothesis 
rescoring~\cite{kwon2025whisfusion, wang2025audio}.

\paragraph{Uniform-state diffusion model.}
USDM works similarly to MDLM but uses a different corruption strategy.

During the forward process,  
tokens are replaced with random samples from the vocabulary rather than a mask token. 
The marginal distribution is defined as $q(z_t \mid w) = \mathrm{Cat}(z_t;\, \alpha_t\, 1_w + (1 - \alpha_t)\boldsymbol{\pi})$, 
where $\boldsymbol{\pi} = \frac{1}{|V|}{1}$ is the uniform distribution over the vocabulary $V$ \cite{austin2021structured}.

During the reverse process, 
these forward dynamics allow for continual token updates. 
Since corrupted tokens are indistinguishable from clean ones, 
the model must re-evaluate every position at each denoising step, 
enabling a self-correcting property where previously predicted tokens 
can be updated or fixed~\cite{sahoo2025diffusion,deschenaux2026diffusion, von2025scaling, sahoo2026scalingmaskeddiffusionlanguage}. 
Consequently, at each denoising step,
the model $w_\theta(z_t, t)$ produces a full probability distribution over the entire vocabulary for every token in the sequence, 
regardless of its current state.
For ASR, 
this dense output is particularly advantageous as it provides a continuous stream of vocabulary-wide probabilities 
that can be directly aligned and combined with frame-wise CTC scores during joint decoding.

 \section{Methodology}
 \label{sec:methodology}

\subsection{Rescoring}
We rescore $n$-best CTC hypotheses $\tilde{a}_1^{\tilde{S}} = (\tilde{a}_1, \dots, \tilde{a}_{\tilde{S}})$
by combining the CTC log-probability with a diffusion language model (DiffLM) score and a prior correction term:
\begin{align}
  S(\tilde{a}_1^{\tilde{S}}) &= \lambda_{\mathrm{CTC}}\log P_{\mathrm{CTC}}(\tilde{a}_1^{\tilde{S}} \mid x_{1}^T) \nonumber \\
  &\quad + \lambda_{\mathrm{DiffLM}} S_{\mathrm{DiffLM}}(\tilde{a}_1^{\tilde{S}}) \nonumber \\
  &\quad - \lambda_{\mathrm{prior}}\log P_{\mathrm{prior}}(\tilde{a}_1^{\tilde{S}})
  \label{eq:rescoring_linear_combination}
\end{align}
where $x_{1}^T$ denotes the sequence of $T$ acoustic feature frames,
$\lambda_{\mathrm{CTC}}$, $\lambda_{\mathrm{DiffLM}}$, and $\lambda_{\mathrm{prior}}$
are tunable interpolation weights,
and the prior term compensates for the implicit language model bias learned by the CTC model \cite{Zeyer:854556, 9383515}.
Since $\log P_{\mathrm{DiffLM}}(\tilde{a}_1^{\tilde{S}})$ is intractable, 
prior work has shown that it can be approximated using a variational 
evidence lower bound (ELBO) \cite{kingma2021on, sahoo2024simple}.
We also propose alternative score functions for the diffusion LM, 
denoted $S_{\mathrm{DiffLM}}(\tilde{a}_1^{\tilde{S}})$.

\paragraph{MDLM.}

A naive sequence-length normalization estimate averages over the full sequence length $\tilde{S}$:
\begin{equation}
    S_{\mathrm{DiffLM}}(\tilde{a}_1^{\tilde{S}}) = - \frac{1}{K} \sum_{k=1}^{K} \frac{1}{\tilde{S}} \frac{\alpha'_{t_k}}{1-\alpha_{t_k}} \sum_{j \in \mathcal{M}_k} \log P_\theta(\tilde{a}_{j} \mid z_{t,k})
\end{equation}
where $t_k$ is a noise level for sample $k$ that determines the masking probability,
$\mathcal{M}_k$ is the set of randomly masked positions in the $k$-th sample,
$\tilde{a}_j$ is the $j$-th token of the hypothesis,
and $z_{t,k}$ denotes the noisy sequence obtained by masking positions $\mathcal{M}_k$ with masking probability $(1 - \alpha_t)$.
As discussed in \cite{nie2025large}, this estimator
suffers from high variance because the weight $\alpha'_{t_k}/(1-\alpha_{t_k})$ 
does not directly reflect the number of masked tokens $|\mathcal{M}_k|$ 
in each Monte Carlo sample, since tokens are masked randomly.

\paragraph{Sample-level mask normalization.}
We normalize each sample by its own mask count before averaging over $K$:
\begin{equation}
    S_{\mathrm{DiffLM}}(\tilde{a}_1^{\tilde{S}}) = \frac{1}{K} \sum_{k=1}^{K} \frac{1}{|\mathcal{M}_k|} \sum_{j \in \mathcal{M}_k} \log P_\theta(\tilde{a}_{j} \mid z_{t,k})
\end{equation}

\paragraph{Global mask normalization.}
We pool all masked predictions across all $K$ samples
and divide by the total mask count:
\begin{equation}
  S_{\mathrm{DiffLM}}(\tilde{a}_1^{\tilde{S}}) = \frac{\sum_{k=1}^{K} \sum_{j \in \mathcal{M}_k} \log P_\theta(\tilde{a}_{j} \mid z_{t,k})}{\sum_{k=1}^{K} |\mathcal{M}_k|}
\end{equation}
Sample-level normalization weights each Monte Carlo sample equally,
whereas global normalization weights each predicted token equally.
As shown in \Cref{fig:mdlm_duo_5ep_rescoring},
they significantly improve WER over sequence-length normalization.

\paragraph{Coupled scoring.}
Additionally, inspired by the coupled-sampling scheme from~\cite{gong2025diffucoder},
we construct $K$ pairs of complementary masks
$\mathcal{M}_k^{(1)}$ and $\mathcal{M}_k^{(2)} = \overline{\mathcal{M}_k^{(1)}}$,
such that every token is masked in exactly one of the two forward passes per pair:
\begin{align}
  &S_{\mathrm{DiffLM}}(\tilde{a}_1^{\tilde{S}}) \nonumber = \frac{1}{K}\sum_{k=1}^{K} \frac{1}{\tilde{S}}
  \Biggl( \sum_{j \in \mathcal{M}_k^{(1)}} \log P_\theta(\tilde{a}_{j} \mid z_{t,k}^{(1)}) \nonumber \\
  &\quad\quad\quad\quad\quad\quad + \sum_{j \in \mathcal{M}_k^{(2)}} \log P_\theta(\tilde{a}_{j} \mid z_{t,k}^{(2)}) \Biggr)
\end{align}
This guarantees that every token contributes to the score.

\paragraph{USDM.}
We adopt the ELBO from~\cite{sahoo2025diffusion} as the scoring objective.
Unlike MDLM, where only masked tokens contribute to the score, 
USDM corrupts positions with uniform noise 
and every token participates in the score regardless of the noise level.

\subsection{Joint-Decoding}

\begin{figure}
  \centering
  \includegraphics[width=\linewidth]{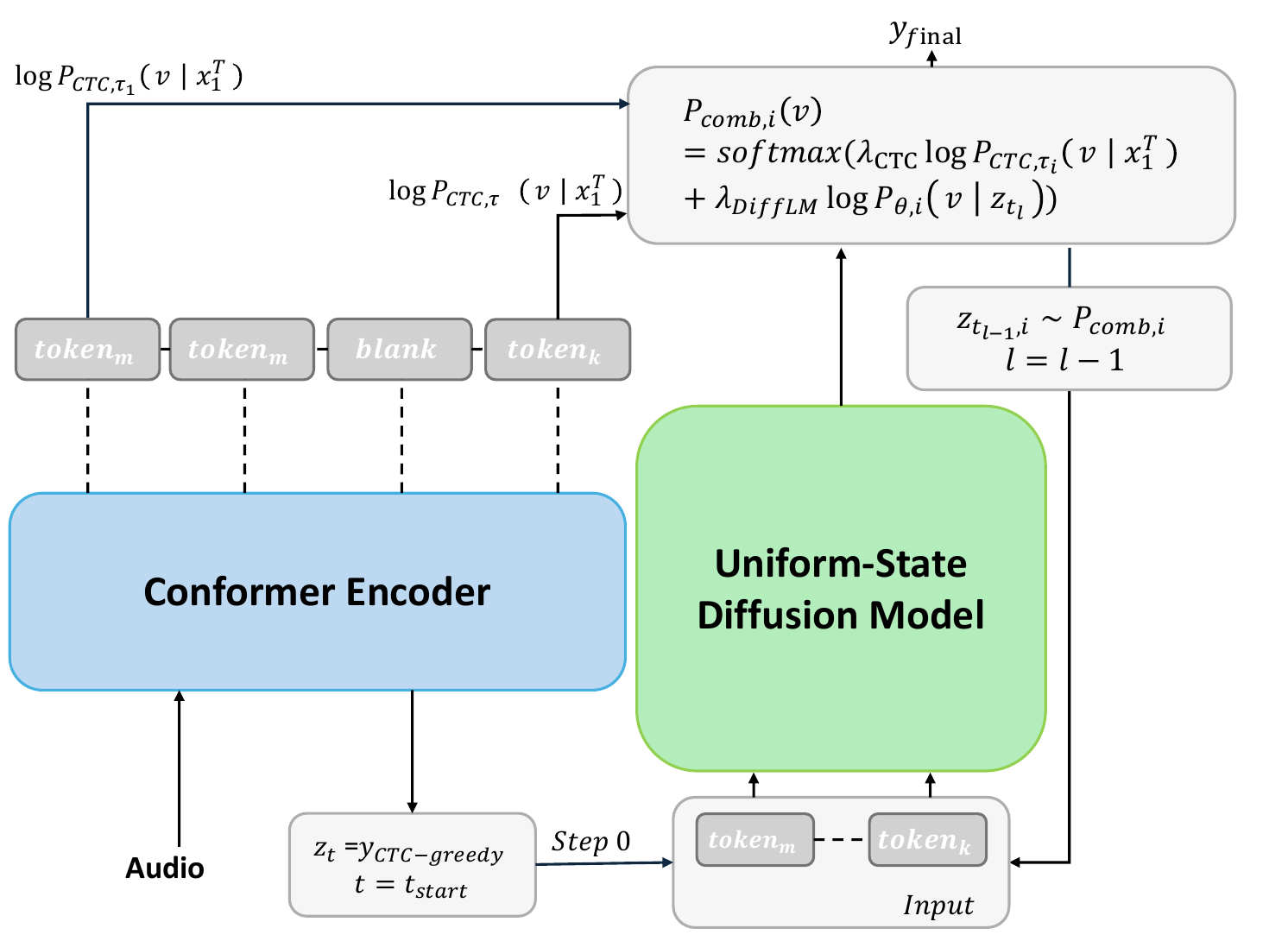}
  \caption{Overview of the proposed joint CTC-USDM decoding. 
  At each denoising step,
  the USDM token-level distribution is combined with the CTC frame-level distribution to sample input to the next denoising step.}
  \label{fig:joint_decoding}
\end{figure}

USDM exhibits several properties that make it particularly suitable for integration with CTC
in a joint decoding framework (see \Cref{fig:joint_decoding}):
its self-correcting nature allows the model to continuously refine all positions rather than committing to predictions early;
and since tokens are corrupted with uniform noise rather than a mask token,
the model maintains a well-defined probability distribution over the full vocabulary at every position
throughout the entire denoising process.
These properties motivate us to extend USDM to a novel joint CTC-USDM decoding framework.

We initialize the denoising process from the CTC greedy sequence at noise level $t_\text{start}$ \cite{tian2026dllmasrfasterdiffusionllmbased},
which we treat as a tunable hyperparameter.
Since CTC operates on frames while USDM operates on tokens,
we align the two by extracting the log-probability distribution of the first frame
corresponding to each collapsed token and renormalizing it over the non-blank vocabulary.
At each denoising step $l$,
and each token position $i$,
USDM produces a token-level distribution $P_\theta(\cdot \mid z_{t_l})$,
which we combine with the CTC distribution:
\begin{align}
  S_{\mathrm{comb}, i}(v) &=
  \lambda_{\mathrm{CTC}} \log P_{\mathrm{CTC}, \tau_i}(v \mid x_1^T) \nonumber \\
  &\quad + \lambda_{\mathrm{DiffLM}} \log P_{\theta, i}(v \mid z_{t_l}),
  \quad \forall v \in V
\end{align}
$\tau_i$ denotes the first CTC frame aligned with token position $i$ in the collapsed greedy sequence.
The term $P_{\mathrm{CTC},\tau_i}(v \mid x_1^T)$ corresponds to the frame-level CTC probability of token $v$ at frame $\tau_i$, obtained from the encoder output and renormalized over the non-blank vocabulary.
The term $P_{\theta,i}(v \mid z_{t_l})$ denotes the token-level probability predicted by the USDM at position $i$ given the current noisy sequence $z_{t_l}$.
The combined score $S_{\mathrm{comb},i}(v)$ therefore integrates acoustic evidence from the CTC model with contextual information from the diffusion language model during each denoising step.

We adopt ancestral sampling from \cite{deschenaux2026diffusion, campbell2022a, austin2021structured},
drawing each position of $z_{t_{l-1}}$ independently from the corresponding combined distribution $P_{\mathrm{comb}, i}$,
which is derived as $P_{\mathrm{comb}, i} = \mathrm{softmax}(S_{\mathrm{comb}, i})$.
\section{Experiments}

\subsection{Experimental Setup}
We trained MDLM and USDM on a combined corpus of normalized LibriSpeech LM data and train-other transcriptions \cite{7178964}. 
For our experiments, 
we leveraged the training frameworks from~\cite{sahoo2024simple, sahoo2025diffusion}.
Models were trained for 5, 10 and 25 epochs using AdamW (0.1 weight decay) \cite{loshchilov2018decoupled}, 
a piecewise linear LR scheduler, 
and a 20,000 token batch size.
Our primary "medium" architecture for both models is a 24-layer 
Diffusion Transformer (DiT) \cite{10377858} with 16 attention heads, 
0.1 dropout \cite{JMLR:v15:srivastava14a} and a 1024-dimensional hidden state.
Text was tokenized via SentencePiece into 10,240 subwords \cite{kudo-richardson-2018-sentencepiece}.

\subsection{Results}

\begin{figure}
  \centering
  \includegraphics[width=\columnwidth]{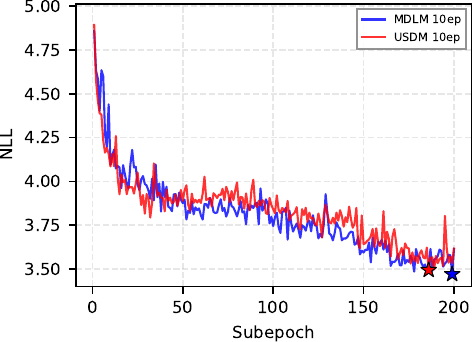}
  \caption{Dev PPL learning curves for MDLM and USDM trained for 10 full epochs (1 full epoch = 20 subepochs).}
  \label{fig:training_curves_ppl_usdm_mdlm_10ep}
\end{figure}

\paragraph{Language model training.}
\Cref{tab:ppl_results_usdm_mdlm_5ep_and_10ep} shows the perplexity upper bounds 
for USDM and MDLM trained with the same configuration.
MDLM achieves lower PPL at 5 and 10 epochs (see \Cref{fig:training_curves_ppl_usdm_mdlm_10ep}, 
e.g.\ 37.0 vs.\ 39.4 on dev at 10 epochs), 
but USDM surpasses it at 25 epochs (34.0 vs.\ 32.3 on dev).
This can be explained by the fact that USDM corrupts tokens with uniform noise rather than explicit mask tokens,
making the task inherently harder since the model must evaluate every position.
Both models show improvement with longer training.


\begin{table}
  \centering
  \caption{Perplexity (PPL) upper bounds on train, dev, and devtrain splits for MDLM and USDM trained for 5, 10 and 25 epochs on LibriSpeech LM data (last checkpoint).}
  \label{tab:ppl_results_usdm_mdlm_5ep_and_10ep}
  \begin{adjustbox}{max width=\linewidth}
    \begin{tabular}{| l | c | c | c | c |}
      \hline
      \multirow{2}{*}{Diffusion LM} & \multirow{2}{*}{Number of Epochs} & \multicolumn{3}{c|}{PPL}\\\cline{3-5}
      & & train & dev & devtrain \\\hline\hline
      \multirow{3}{*}{MDLM} & 5 & $\leq$ 36.2 & $\leq$ 36.6 &  $\leq$  47.9\\\cline{2-5}
       & 10 & $\leq$ 32.7 & $\leq$ 37.0 &  $\leq$  44.6\\\cline{2-5}
       & 25 & $\leq$ 30.0 & $\leq$ 34.0 &  $\leq$  42.2 \\\hline\hline
       \multirow{3}{*}{USDM} & 5 & $\leq$ 40.5 & $\leq$ 40.2 &  $\leq$  59.5\\\cline{2-5}
       & 10 & $\leq$ 37.0 & $\leq$ 39.4 &  $\leq$  48.0 \\\cline{2-5}
       & 25 & $\leq$ 33.5 & $\leq$ 32.3 &  $\leq$  44.1 \\\hline
    \end{tabular}
\end{adjustbox}
\end{table}

\begin{figure}
  \centering
  \includegraphics[width=\columnwidth]{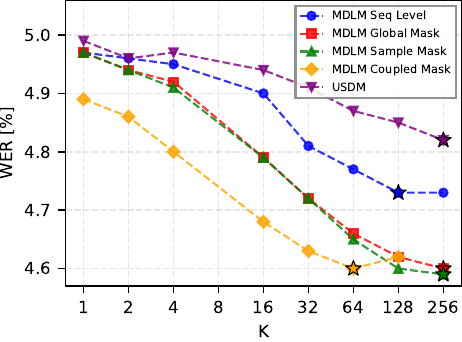}
  \caption{WER [\%] on dev-other comparing MDLM rescoring 
  (sequence-level, global-mask and sample-mask score normalization), 
  MDLM (5 ep) coupled scoring, 
  and USDM (5 ep) rescoring, 
  across different numbers of Monte Carlo samples ($K$). 
  Stars mark the best WER for each method.}
  \label{fig:mdlm_duo_5ep_rescoring}
\end{figure}

\paragraph{Rescoring.}
\Cref{fig:mdlm_duo_5ep_rescoring} compares MDLM and USDM rescoring strategies across varying numbers of Monte Carlo samples $K$.
As illustrated in \Cref{fig:mdlm_duo_5ep_rescoring}, 
MDLM rescoring consistently outperforms the CTC baseline (5.08\% WER) and USDM. 
While standard sequence-length normalization achieves a WER of 4.73\% ($K=256$),
our proposed sample-level mask normalization yields a further reduction to 4.59\%. 
For all rescoring experiments,
we do not sample but fix the noise level to $t = 0.5$.
In future work,
we will investigate other noise schedules and their influence on the results.
Additionally, 
coupled scoring improves results with fewer samples, 
reaching 4.60\% WER at $K=64$.
Furthermore, as shown in \Cref{tab:mdlm_5ep_10ep_25ep_comparison}, 
extending MDLM training to 10 and 25 epochs provides additional gains, 
reaching a rescoring performance of 4.56\% and 4.52\% WER at $K=256$.
While USDM rescoring does not match the performance of MDLM, 
it still yields improvements over the CTC baseline. 
As shown in \Cref{tab:usdm_5ep_10ep_25ep_comparison}, USDM reduces the WER from 5.08\% to 4.82\% at $K=256$.
With an increased number of trained epochs, the WER decreased to 4.80\% for 25 epochs.

\begin{table}
  \centering
  \caption{MDLM rescoring (with sample-level normalization) WER [\%] on dev-other across 5, 10 and 25 training epochs, 
  varying the number of Monte Carlo samples ($K$). Standard deviations computed over 5 random seeds.}
  \label{tab:mdlm_5ep_10ep_25ep_comparison}
  \begin{adjustbox}{max width=\linewidth}
    \begin{tabular}{| l | c | c | c |}
    \hline
    \multirow{2}{*}{$K$} & \multicolumn{3}{ c |}{WER [\%]}\\\cline{2-4}
     & MDLM 5 ep & MDLM 10 ep & MDLM 25 ep\\
    \hline\hline
    1   & 4.97 $\pm$ .03 & 4.94 $\pm$ .01 & 4.95 $\pm$ .01 \\ \hline
    2   & 4.94 $\pm$ .01 & 4.94 $\pm$ .02 & 4.91 $\pm$ .02 \\ \hline
    16  & 4.79 $\pm$ .03 & 4.78 $\pm$ .02 & 4.75 $\pm$ .03 \\ \hline
    32  & 4.72 $\pm$ .03 & 4.67 $\pm$ .02 & 4.65 $\pm$ .03 \\ \hline
    64  & 4.65 $\pm$ .04 & 4.60 $\pm$ .02 & 4.56 $\pm$ .02 \\ \hline
    128 & 4.60 $\pm$ .02 & 4.58 $\pm$ .02 & 4.55 $\pm$ .02 \\ \hline
    256 & 4.59 $\pm$ .02 & 4.56 $\pm$ .03 & 4.52 $\pm$ .01 \\ \hline
    \end{tabular}
  \end{adjustbox}
\end{table}

\begin{table}
  \centering
  \caption{USDM rescoring WER [\%] on dev-other across 5, 10 and 25 training epochs, 
  varying the number of Monte Carlo samples ($K$). Standard deviations computed over 5 random seeds.}
  \label{tab:usdm_5ep_10ep_25ep_comparison}
  \begin{adjustbox}{max width=\linewidth}
    \begin{tabular}{| l | c | c | c |}
    \hline
    \multirow{2}{*}{$K$} & \multicolumn{3}{ c |}{WER [\%]}\\\cline{2-4}
     & USDM 5 ep & USDM 10 ep & USDM 25 ep\\
    \hline\hline
    1   & 4.99 $\pm$ .03 & 4.97 $\pm$ .02 & 4.97 $\pm$ .01 \\ \hline
    2   & 4.96 $\pm$ .03 & 4.99 $\pm$ .03 & 4.99 $\pm$ .02 \\ \hline
    16  & 4.94 $\pm$ .03 & 4.92 $\pm$ .02 & 4.92 $\pm$ .02 \\ \hline
    32  & 4.91 $\pm$ .02 & 4.89 $\pm$ .04 & 4.89 $\pm$ .03 \\ \hline
    64  & 4.87 $\pm$ .04 & 4.85 $\pm$ .04 & 4.86 $\pm$ .02 \\ \hline
    128 & 4.85 $\pm$ .03 & 4.83 $\pm$ .02 & 4.80 $\pm$ .01 \\ \hline
    256 & 4.82 $\pm$ .02 & 4.82 $\pm$ .02 & 4.80 $\pm$ .02\\ \hline
    \end{tabular}
  \end{adjustbox}
\end{table}

\paragraph{CTC-USDM Joint Decoding.}
As shown in \Cref{tab:ctc_usdm_5_ep_joint_different_t}, 
joint decoding yields better results than USDM rescoring, 
suggesting that the active participation of USDM in hypothesis construction produces better hypotheses.
A lower initial noise level ($t_{\text{start}} = 0.3$) allows the model to reach optimal WER with fewer denoising steps, 
while all configurations converge to the same final WER.
As shown in \Cref{tab:ctc_usdm_5_10_25_ep_joint_03_t}, 
extending USDM training to 10 and 25 epochs further improves joint decoding, 
reaching a peak WER of 4.73\% and 4.71\%, respectively.

\begin{table}
  \centering
  \caption{WER [\%] on dev-other for CTC+USDM joint decoding with different initial noise level $t_{\text{start}}$, and denoising steps $L$ (for all experiments, $\lambda_{\mathrm{DiffLM}} = 0.3$ shows the best performance).}
  \label{tab:ctc_usdm_5_ep_joint_different_t}
  \begin{adjustbox}{max width=\linewidth}
    \begin{tabular}{| l || c | c | c | c | c | c | c |}
    \hline
    & \multicolumn{7}{c|}{WER [\%]} \\\cline{2-8}
    & \multicolumn{7}{c|}{$L$} \\\cline{2-8}
    $t_{\text{start}}$ & 1 & 8 & 12 & 16 & 32 & 48 & 64 \\\hline\hline
    0.3 & 4.79 & 4.78 & 4.79 & 4.77& 4.78 & 4.77 & 4.77 \\\hline
    0.5 & 4.81 & 4.81 & 4.80 & 4.78 & 4.77 & 4.77 & 4.77 \\\hline
    0.8 & 4.82 & 4.79 & 4.78 & 4.78 & 4.77 & 4.77 & 4.77 \\\hline
    \end{tabular}
  \end{adjustbox}
\end{table}

\begin{table}
  \centering
  \caption{WER [\%] on dev-other for CTC+USDM joint decoding comparing models trained on different numbers of epochs with $\lambda_{\mathrm{DiffLM}} = 0.3$ and $t_{\text{start}} = 0.3$.}
  \label{tab:ctc_usdm_5_10_25_ep_joint_03_t}
  \begin{adjustbox}{max width=\linewidth}
    \begin{tabular}{| l || c | c | c | c | c | c | c |}
    \hline
    & \multicolumn{7}{c|}{WER [\%]} \\\cline{2-8}
    & \multicolumn{7}{c|}{$L$} \\\cline{2-8}
    Epochs & 1 & 8 & 12 & 16 & 32 & 48 & 64 \\\hline\hline
    5 & 4.79 & 4.78 & 4.79 & 4.77& 4.78 & 4.77 & 4.77 \\\hline
    10 & 4.74 & 4.73 & 4.74 & 4.76 & 4.74 & 4.74 & 4.73 \\\hline
    25 & 4.74 & 4.72 & 4.74 & 4.72 & 4.73 & 4.72 & 4.71 \\\hline
    \end{tabular}
  \end{adjustbox}
\end{table}

\paragraph{Comparison with Autoregressive LMs.}
\Cref{tab:ctc_all_models_5ep_rescoring_comp} provides an overview of the results across the different modeling approaches. 
As expected, 
the autoregressive language models achieve the lowest overall WER, 
with 4.19\% for rescoring and 3.86\% for joint decoding. 
Additionally, 
for the autoregressive model, increasing the number of layers from 12 to 24 reduces rescoring WER from 4.29\% to 4.19\%.
While we do not investigate scaling of diffusion LMs further, some prior works on denoising language models \cite{koch2025reproducingdissectingdenoisinglanguage, gu2026revisitingasrerrorcorrection} 
and diffusion language models \cite{prabhudesai2025diffusion} investigate their scaling behavior and
show that they surpass autoregressive models under data-constrained settings 
if trained long enough. 
Motivated by this, 
in future work we plan to compare diffusion and autoregressive LMs at larger model and data scales.

\begin{table}
  \centering
  \caption{WER [\%] on dev-other comparing CTC without LM, 
  with an autoregressive LM, with MDLM (10 ep, sample-level mask norm., $K=256$) 
  and USDM (10 ep, $K=256$ rescoring and $L=64$ joint-decoding).}
  \label{tab:ctc_all_models_5ep_rescoring_comp}
  \begin{adjustbox}{max width=\linewidth}
    \setlength{\tabcolsep}{5pt}
    \begin{tabular}{| l | c | c | c | c |}
    \hline
    \multirow{2}{*}{LM} & \multirow{2}{*}{\makecell{Model\\Dim}} & \multirow{2}{*}{\makecell{Num\\Layers}} & \multicolumn{2}{c|}{WER [\%]} \\\cline{4-5}
    & & &\multicolumn{2}{c|}{dev-other}   \\\hline\hline
    None & -- & -- & greedy & 5.08\\\hline\hline
    \multirow{4}{*}{\makecell[l]{Auto-\\regressive\\LM}} & \multirow{2}{*}{768} & \multirow{2}{*}{12} & rescoring & 4.29 \\\cline{4-5}
     & & & joint-decoding & 4.10\\\cline{2-5}
     & \multirow{2}{*}{1024} & \multirow{2}{*}{24}& rescoring & 4.19\\\cline{4-5}
     & & & joint-decoding & 3.86\\\hline\hline
    MDLM & 1024 & 24 & rescoring & 4.56\\\hline\hline
    \multirow{2}{*}{USDM} & \multirow{2}{*}{1024} & \multirow{2}{*}{24} & rescoring & 4.82\\\cline{4-5}
     & & & joint-decoding & 4.73\\\hline
    \end{tabular}
  \end{adjustbox}
\end{table}

\section{Conclusions}

In this work, 
we systematically explored the integration of discrete diffusion language models into ASR systems. 
While traditional autoregressive models are constrained by a strictly sequential, 
left-to-right decoding structure, 
diffusion LMs leverage bidirectional context and parallel generation, 
offering a more flexible and theoretically faster alternative for ASR.
We introduced new methods to rescore ASR hypotheses using MDLM, 
namely Global and Sample-level Mask Normalization. 
By utilizing the mask length for normalization, 
these methods significantly improved performance in comparison to standard sequence-level normalization.
Most notably, we noticed unique properties of Uniform-State Diffusion Models, 
specifically their lack of artificial mask tokens and their full-vocabulary probability distribution for each position, 
and developed a CTC-USDM joint decoding framework, 
which successfully outperformed static rescoring with USDM.
Evaluation shows MDLM achieves better rescoring accuracy on limited data than USDM. 
This is likely because MDLM's explicit mask tokens provide a clearer reconstruction signal, 
whereas USDM's uniform noise forces the model to implicitly distinguish clean from noisy tokens at every position. 
Consequently, USDM's more complex objective may require additional data or scaling to reach peak performance.
We also compared our methods with standard autoregressive language models.
Autoregressive LMs show better performance in both rescoring and joint decoding,
however, prior studies suggest that diffusion language models can surpass autoregressive models when trained long enough.
In future work, 
we plan to evaluate these models with more training epochs and scale the model capacities further to close the performance gap with autoregressive baselines.
We also plan to investigate the joint decoding framework in greater detail, including extending it to MDLM.
\ifcameraready
\section{Acknowledgements}

This work was partially supported by NeuroSys,
which as part of the initiative “Clusters4Future” is funded by the Federal Ministry of
Education and Research BMBF (funding IDs 03ZU2106DA and 03ZU2106DD),
and by the project RESCALE within the program
\textit{AI Lighthouse Projects for the Environment, Climate, Nature and Resources}
funded by
the Federal Ministry for the Environment, Nature Conservation,
Nuclear Safety and Consumer Protection (BMUV),
funding ID: 67KI32006A.
The authors gratefully acknowledge the computing time provided to them
at the NHR Center NHR4CES at RWTH Aachen University
(project number p0023565 and p0023999).
This is funded by the Federal Ministry of Education and Research,
and the state governments participating on the basis
of the resolutions of the GWK for
national high performance computing at universities
(\url{www.nhr-verein.de/unsere-partner}).
\fi

\section{Generative AI Use Disclosure}
We use LLMs to improve the formulations and grammar of the paper.

\bibliographystyle{IEEEtran}
\bibliography{mybib}

\end{document}